\documentclass[10pt,twocolumn,letterpaper]{article}

\usepackage[pagenumbers]{cvpr} 
\usepackage{graphicx}
\usepackage{amsmath}
\usepackage{amssymb}
\usepackage{booktabs}

\usepackage{times}
\usepackage{epsfig}
\usepackage{graphicx}
\usepackage{amsthm}
\usepackage{caption}
\usepackage{multirow}
\usepackage{makecell}
\usepackage{comment}
\usepackage[table,xcdraw,dvipsnames]{xcolor}
\usepackage{algorithm}
\usepackage{algpseudocode}

\usepackage[symbol]{footmisc}

\usepackage[pagebackref,breaklinks,colorlinks]{hyperref}

\usepackage[capitalize]{cleveref}
\crefname{section}{Sec.}{Secs.}
\Crefname{section}{Section}{Sections}
\Crefname{table}{Table}{Tables}
\crefname{table}{Tab.}{Tabs.}

\begin{document}

\title{Slot-VPS: Object-centric Representation Learning for Video Panoptic Segmentation}

\author{First Author\\
Institution1\\
Institution1 address\\
{\tt\small firstauthor@i1.org}
\and
Second Author\\
Institution2\\
First line of institution2 address\\
{\tt\small secondauthor@i2.org}
}

\author {Yi Zhou\textsuperscript{\rm1}, 
Hui Zhang\textsuperscript{\rm1},
Hana Lee\textsuperscript{\rm2},
Shuyang Sun\textsuperscript{\rm 3},
Pingjun Li\textsuperscript{\rm1},
Yangguang Zhu\textsuperscript{\rm1}, \\
ByungIn Yoo\textsuperscript{\rm2},
Xiaojuan Qi\textsuperscript{\rm4\footnotemark[1]} ,
Jae-Joon Han\textsuperscript{\rm2\footnotemark[1]} \\
 \textsuperscript{\rm 1}Samsung Research China - Beijing (SRC-B) \\
 \textsuperscript{\rm 2}Samsung Advanced Institute of Technology (SAIT), South Korea \\
 \textsuperscript{\rm 3}University of Oxford
 \textsuperscript{\rm 4}The University of Hong Kong\\
 {\tt\small \{yi0813.zhou, hui123.zhang, hana.hn.lee, byungin.yoo, jae-joon.han\}@samsung.com}\\
 {\tt\small kevinsun@robots.ox.ac.uk}, 
 {\tt\small xjqi@eee.hku.hk}
 }
\maketitle

\renewcommand{\thefootnote}{\fnsymbol{footnote}}
\footnotetext[1]{Corresponding author.}

\begin{abstract}
Video Panoptic Segmentation (VPS) aims at assigning a class label to each pixel, uniquely segmenting and identifying all object instances consistently 
across all frames.
Classic solutions usually decompose the VPS task into several sub-tasks and utilize multiple surrogates (\eg boxes and  masks, centers and offsets) to represent objects. 
However, 
this divide-and-conquer strategy 
requires complex post-processing in both spatial and temporal domains and is vulnerable to failures from surrogate tasks.
In this paper,
inspired by object-centric learning which learns compact and robust object representations, 
we present Slot-VPS, 
the first end-to-end framework for this task.
We  
encode all panoptic entities in a video, including both foreground instances and background semantics, with a unified representation called panoptic slots.
The coherent spatio-temporal object's information is retrieved and encoded into the panoptic slots by the proposed Video Panoptic Retriever,
enabling to localize, segment, differentiate, and associate objects in a unified manner.
Finally, the output panoptic slots can be directly converted into the class, mask, and object ID of panoptic objects in the video. 
We conduct extensive ablation studies and demonstrate the effectiveness of our approach on two benchmark datasets, Cityscapes-VPS (\textit{val} and test sets) and VIPER (\textit{val} set), achieving new state-of-the-art performance of $63.7, 63.3$ and $56.2$ VPQ, respectively.
\end{abstract}

\section{Introduction}
\label{sec:intro}
Video panoptic segmentation (VPS) \cite{kim2020video, woo2021learning, qiao2020vip} aims at classifying all foreground instances (\emph{things}), \eg cars, people, \etc, and countless background semantics (\emph{stuff}), \eg sky, road, \etc 
, segmenting and tracking all object instances consistently across all frames. It is beneficial to many high-level video understanding tasks, such as Video Question Answering \cite{park2021bridge} and Video Captioning \cite{zheng2020syntax}, and various real-world applications, such as autonomous driving and video editing.

\begin{figure}[t]
\centering
\includegraphics[width=\linewidth]{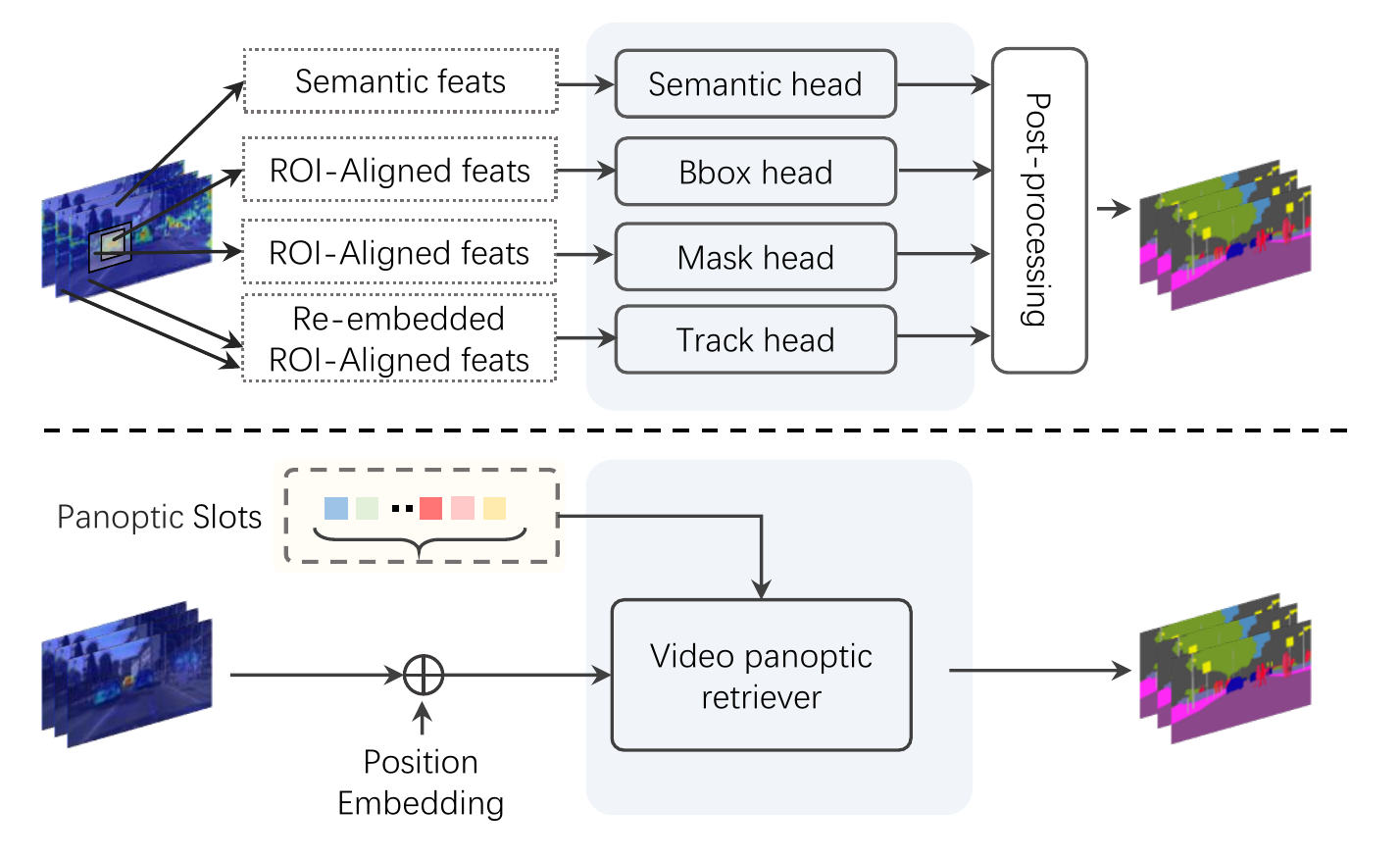}
\caption{\textbf{Comparison between previous works \cite{kim2020video, voigtlaender2020siam} and the proposed Slot-VPS.}
VPSNet represents objects with multiple representations, relies on several sub-networks,
and requires complex post-processing (\eg NMS, things-stuff fusion, similarity score fusion for tracking),
while we introduce panoptic slots to uniformly represent panoptic objects 
(\ie things and stuff) in a video, enabling a unified end-to-end framework.
}
\label{fig:front_figure}
\end{figure}

Existing methods \cite{kim2020video, woo2021learning, qiao2020vip} model things and stuff in a video separately with several sub-networks (as shown in Figure \ref{fig:front_figure}) tailored to different sub-tasks including semantic segmentation \cite{long2015fully,chen2017deeplab},
instance segmentation \cite{he2017mask,bolya2019yolact, lee2020centermask}, and tracking \cite{wang2020towards, sun2020transtrack}. 
Complicated post-processing in both spatial (\eg things-stuff fusion) and temporal (\eg similarity scores fusion for instance association) domains is needed to fuse predictions of different subtasks into final VPS results.
However, such a decomposed pipeline suffers from several issues.
First, the complicated post-processing is time-consuming and requires manual parameter tuning, which is likely to produce sub-optimal results.
Second, erroneous predictions from different branches will adversely affect each other and harm the overall performance.
For example, the inaccurate boxes will also lead to  incomplete segmentation masks, and missing centers will deteriorate the temporal tracking results,  
which can barely be corrected by post-processing.
Third, end-to-end training is blocked, potentially hindering the model from learning features directly optimized for the VPS task.

To solve the aforementioned problems,
motivated by object-centric representation learning which learns the compact and robust representations of objects, we introduce a unified end-to-end framework, Slot-VPS, as illustrated in Figure \ref{fig:front_figure}.
All \emph{panoptic objects} (including both stuff and things) in the video are represented as a unified representation named \emph{panoptic slots}.
Panoptic slots are a set of learnable parameters and can be updated through interacting with features extracted from videos.
Each panoptic slot is responsible for a stuff class or an object instance in the video, enabling the direct predictions of the class, mask, and object ID of each panoptic object in an end-to-end fashion.

To encode the spatio-temporal information of video-level panoptic objects 
into the panoptic slots,
we introduce the
Video Panoptic Retriever (VPR).
VPR incorporates a Panoptic Retriever to retrieve the location and appearance information from the spatial features for panoptic object localization and segmentation, and a Video Retriever to correlate slots across different time steps for temporally associating object instances.
Furthermore, 
during the above process,
softmax-based operation, which normalizes the contributing weights of each slot, is performed to encourage the panoptic slots to compete and be distinct from each other such that the redundancy among slots will be suppressed.
Finally, the spatio-temporal coherent panoptic slots, carrying both object's spatial information and temporal identification information, can be utilized to directly predict final results, \ie class, mask, and object ID of panoptic objects in the video.

To our best knowledge, this is the first fully
unified end-to-end framework for the VPS task.
It does not rely on any surrogates in both spatial and temporal domains and hence bypasses the drawbacks of dependence on complex post-processing and influence from sub tasks' failures.

Experimental results on the Cityscapes-VPS \cite{kim2020video} and VIPER \cite{kim2020video} datasets demonstrate the effectiveness of our method. 
Thanks to the unified end-to-end framework and the object-centric learning, 
our method outperforms the state-of-the-art \cite{kim2020video, qiao2020vip} on the \textit{val} and test sets ($63.7$, $63.3$ VPQ) of Cityscapes-VPS,
and the \textit{val} set ($56.2$ VPQ) of VIPER with better efficiency.

\begin{figure}[t]
\centering
\includegraphics[width=\linewidth]{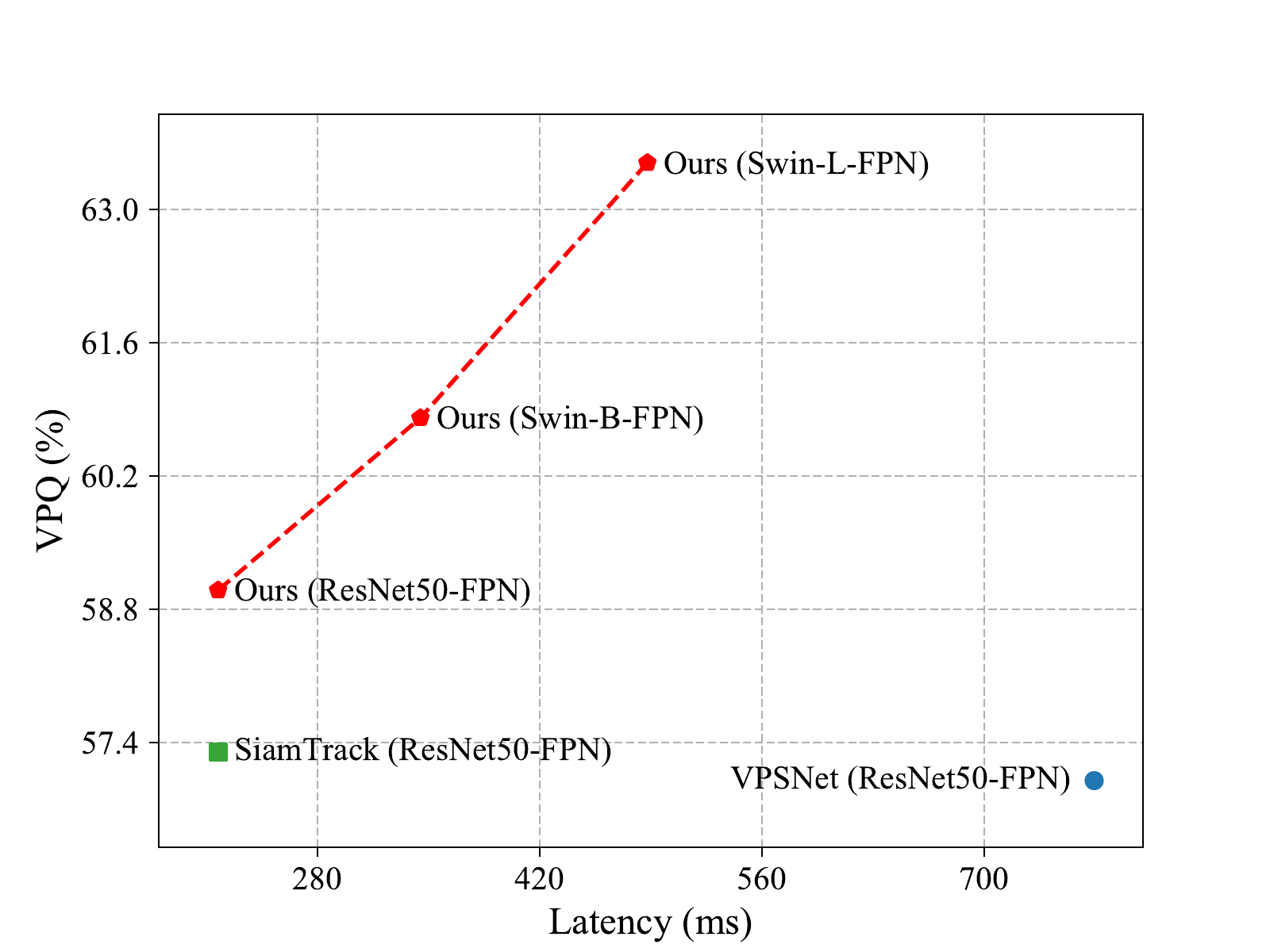}
\caption{\textbf{Speed-Accuracy trade-off curve on the Cityscapes-VPS \textit{val} set.}
The latency is measured on V100 GPU.
}
\label{fig:comparison_fps}
\end{figure}

Our main contributions can be summarized as follows:
\begin{itemize}
    \item
    We propose to uniformly represent all panoptic objects in the video with a unified representation (\ie panoptic slots), and introduce Slot-VPS, 
    the first unified 
    end-to-end pipeline for the VPS task.
    \item 
    To spatially localize, segment, differentiate and temporally associate objects,
    Video Panoptic Retriever (VPR)
    is developed to retrieve and encode spatio-temporal coherent objects' information into panoptic slots.
    \item 
    Our method outperforms the state-of-the-arts \cite{kim2020video, qiao2020vip}
    on both Cityscapes-VPS and VIPER datasets.
    What's more, as shown in Figure \ref{fig:comparison_fps}, our model has better efficiency.
\end{itemize}

\section{Related Work}

\begin{figure*}[t]
\centering
\includegraphics[width=\textwidth]{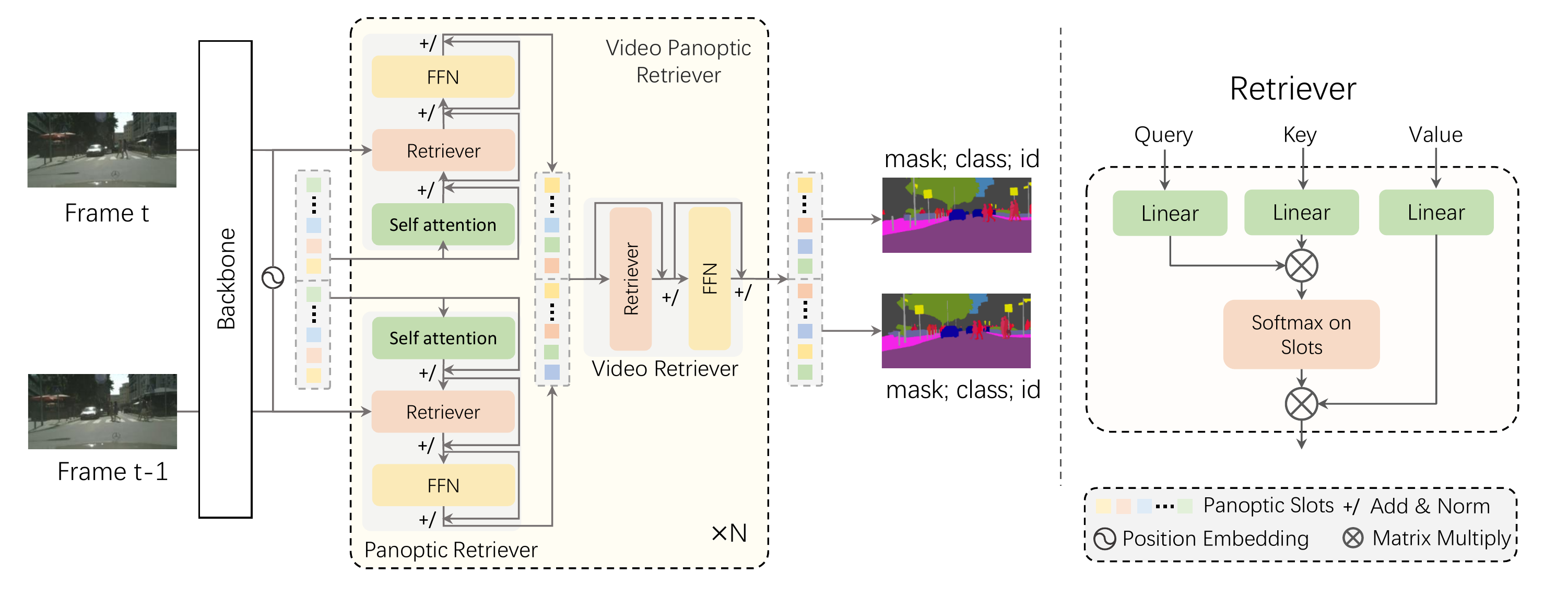}
\caption{\textbf{Overview of the Slot-VPS.}
Take two frames (t and t-1) as an example, 
the multi-scale features extracted from the backbone with position embedding and panoptic slots are fed into Video Panoptic Retriever (VPR) modules for $N$ stages to generate the spatio-temporal coherent panoptic slots.
Panoptic slots, representing the panoptic objects in the video, are shared among all frames initially.
Finally they are directly converted into
objects' masks, classes and IDs.
Note that FFN stands for Feed Forward Network.
}
\label{fig:method}
\end{figure*}

\noindent\textbf{Panoptic Segmentation (PS).}
Unifying semantic and instance segmentation on the image level,
the PS
task \cite{kirillov2019panoptic, PanopticFPN, xiong2019upsnet, li2018weakly, Panoptic-deeplab, lazarow2020learning} requires assigning a class label for all pixels and uniquely segmenting all object instances.
Early attempts \cite{kirillov2019panoptic} in PS follow the decomposition pipeline,
separately predicting the semantic and instance segmentation results and then adopting the things-stuff fusion process in later stages.
Several works 
try to simplify the process and improve the accuracy through
replacing the post things-stuff fusion with the parameter-free \cite{xiong2019upsnet} or trainable \cite{li2020unifying} panoptic head.
Furthermore, more researchers \cite{li2020fully, tian2020conditional} try to abandon the separated branches and
build an end-to-end unified framework.
A recent trend in PS task \cite{carion2020end, wang2021max, cheng2021per} is regarding this task as the set prediction problem and try to construct the concise end-to-end networks with the help of transformers.
However, note that all these unified methods only consider the unification of things and stuff on the spatial domain, excluding the temporal domain.

\noindent\textbf{Video Panoptic Segmentation (VPS).} 
As a direct extension of the PS task into video domain, previous approaches in the VPS task
\cite{kim2020video, qiao2020vip} usually apply the image-level approach \cite{xiong2019upsnet, Panoptic-deeplab} to each video frame and try to associate the result of all frames by using an extra temporal association head (\eg track head \cite{kim2020video, woo2021learning}, temporal center regression head \cite{qiao2020vip}), \etc.
However, all these methods represent objects with multiple representations (\eg boxes and masks, centers and offsets), requiring multiple separate networks to handle sub-tasks of VPS and complex post-processing (\eg NMS, things-stuff fusion, similarity fusion for tracking) 
on both spatial and temporal domains.
To the best of our knowledge, this is the first time that the unification of things and stuff in both spatial and temporal domains are discussed simultaneously for this task.

\noindent\textbf{Transformers.}
Inspired by the success of Transformers \cite{vaswani2017attention} in Natural Language Processing (NLP) tasks, 
numerous researches have also been investigated in the computer vision community,
\eg Object Detection \cite{carion2020end, zhu2020deformable}, 
Panoptic Segmentation \cite{wang2021max, cheng2021per}, and
Video Instance Segmentation \cite{wang2021end, yang2019video, liu2021sg}, \etc.
In these pipelines, objects are represented as a set of vectors containing all related information such as location, appearance, \etc 
Attention in transformers is responsible for localizing and segmenting objects, and the bipartite-matching based mechanisms help separate different objects.
With the help of these technologies,
above tasks can be converted into direct set prediction problem without many hand-designed components.
Different from all these works, we introduce the slots competing mechanism into the learning process to enhance the discriminability of objects in both spatial and temporal domains.
Jointly representing stuff and things on the video level with panoptic slots, we propose the fully unified 
end-to-end
framework in which all operations in both spatial and temporal domains are fulfilled based on the panoptic slots.

\noindent\textbf{Object-centric learning.} 
Object-centric representation learning \cite{gao2016object, engelcke2019genesis, locatello2020object, liu2021semantic, adeli2021recurrent} mainly focuses on learning robust, generalizable object representations in various scenarios such as unsupervised object discovery, novel-viewpoint prediction, \etc 
In \cite{locatello2020object}, Francesco \etal propose the Slot Attention to predict a set of task-dependent abstract representations called \emph{slots} from the perceptual representations such as the output of a convolutional neural network.
Slots are exchangeable and can bind to any object in the input.
Different from the attention in transformer, the Slot Attention lets the randomly sampled slots compete with each other in the iterative learning process, which will further help the discrimination of objects. 
However, the Slot Attention is mainly 
applied to synthetic dataset scenarios,
assuming a normal distribution for the slots.
In this paper, motivated by the object-centric representations and the competing mechanism, we introduce the panoptic slots and the VPR module to successfully
bring object-centric representation learning to real-world data, and set up new state-of-the-art for the VPS task on 
two datasets.

\section{Method}
\subsection{Model Architecture}
As shown in Figure \ref{fig:method}, the Slot-VPS framework consists of a backbone (including Resnet50 \cite{he2016deep}, FPN \cite{lin2017feature}, several deformable convolutions \cite{dai2017deformable, xiong2019upsnet}) for extracting multi-scale features
and Video Panoptic Retriever (VPR) modules for $N$ stages. 
Each stage, containing several VPR modules, is responsible for features of a certain scale.
Given two consecutive frames of an input video as an example,
we denote features of two frames at a certain scale extracted from backbone
as $\mathbf{X_t}, \mathbf{X_{t-1}} \in \mathbb{R}^{D \times C}$
where $D, C, t$ refer to the spatial size (height $\times$ width), the number of channels of feature maps, and time index, respectively.
The VPR takes the certain scale features of two frames ($\mathbf{X_t}$,  $\mathbf{X_{t-1}}$), position embedding ( $\mathbf{P} \in \mathbb{R}^{D \times C}$), and \textit{panoptic slots} ($\mathbf{S} \in \mathbb{R}^{L \times C}$) as inputs to generate spatio-temporal coherent panoptic slots, where $L, C$ denote the number of slots and channels, respectively.
The final prediction heads further utilize the output panoptic slots to predict the classes, masks, and IDs of panoptic objects in the video.
Note that the dimension index of the mini-batch is omitted for clarity.

\noindent\textbf{Panoptic slots.}
To unify the representation in the video, we define panoptic slots, a set of learnable parameters 
$\mathbf{S} \in \mathbb{R}^{L \times C}$, 
to represent all panoptic objects (both \textit{things} and \textit{stuff}) within a video.
Each slot corresponds to an object, hence the slot number $L$  represents the number of possible panoptic objects (\eg $100$) in the video.
Panoptic slots are 
randomly initialized
and can be gradually optimized through interacting with spatio-temporal information. 

\subsection{Video Panoptic Retriever (VPR)}
The VPR consists of the Panoptic Retriever 
and the Video Retriever.
For each module $i \in \{1, ..., U\}$,
where $U$ is the total number of the VPR modules in the network, 
the Panoptic Retriever associates the input panoptic slots 
$\mathbf{S}_t^{i}, \mathbf{S}_{t-1}^{i} \in \mathbb{R}^{L \times C} $ with the features of each frame to produce spatially coherent output panoptic slots 
$\mathbf{\hat S}^{i}_t, \mathbf{\hat S}^{i}_{t-1} \in \mathbb{R}^{L \times C} $ for each frame.
In this process, 
Panoptic Retriever
retrieves the object's information from features through an attention structure called \emph{Retriever}.
Then the Video Retriever, taking the panoptic slots 
$\mathbf{\hat{S}}_t^{i}$ and $\mathbf{\hat{S}}_{t-1}^{i}$ as input,
further utilizes the Retriever to extract temporal correlations between these panoptic slots for temporally enhanced panoptic slots.
The spatio-temporal refined panoptic slots are then forwarded into the next VPR for iterative refinements.
Note that the $\mathbf{S}_t^0$, $\mathbf{S}_{t-1}^0$ are identical to $\mathbf{S}^0$ for the first stage.

\noindent\textbf{Retriever (RE).}
We introduce the Retriever, which is
for retrieving information correlated to the query.
This module can be cast in two different aspects.
For the spatial domain, it is considered as a learning process to map from spatial features to panoptic slots.
For the temporal domain, it can be regarded as learning association between slots of different 
time 
frames.
Different from classic dot-product attention \cite{vaswani2017attention}, we incorporate a slots competing mechanism \cite{locatello2020object} to Retriever, which enables better discrimination of objects, considering that each object should be distinct from other objects both in spatial and temporal domains. 

Here we introduce the 
formulation of Retriever.
Denote the inputs of Retriever as Query $\mathbf{Q} \in \mathbb{R}^{L_{q} \times C}$, Key $\mathbf{K} \in \mathbb{R}^{D_{v} \times C}$ and Value $\mathbf{V} \in \mathbb{R}^{D_{v} \times C}$,
as shown in Figure \ref{fig:method}.
The process of Retriever can be explained by three steps
, including information transformation, correlation calculation, and correlated information retrieval.

Three Linear layers are first applied to transform panoptic slots and target information
into the common space.
We can denote the transformed slots, keys, and values as $\mathbf{Q}_{\theta} \in \mathbb{R}^{ L_{q} \times C}$, $\mathbf{K}_{\phi} \in \mathbb{R}^{D_{v} \times C}$, $\mathbf{V}_{g} \in \mathbb{R}^{D_{v} \times C}$
respectively.

In the second step, the correlation between $L_{q}$ slots and $D_{v}$ vectors is
associated with
matrix multiplication operation,
resulting in a correlation matrix $\mathbf{M} \in \mathbb{R}^{D_k \times L_q}$. 
If the specific object represented by the slot is correlated with a specific vector, then the corresponding value in the correlation matrix will be high.
Furthermore,
to alleviate the phenomenon that two slots correspond to the same target vector, 
we let the slots compete with each other through applying Softmax on the slot dimension.
The above process can be formulated as:
\begin{equation}
\begin{split}
    \mathbf{M} &=
    \mathbf{K}_{\phi} \cdot
    \mathbf{Q}^{T}_{\theta}
    ,
    \\
    \mathbf{A}_{x, y} = \frac{e^{M_{x,y}}}{\sum_{l}{e^{M_{x,l}}}}&,
    \textrm{ for}\ x = 1, ..., D_{v}; y=1, ..., L_{q},
    \\
\end{split}
\label{eq:attn_calculation}
\end{equation}
where $\mathbf{M}_{x, y}$, $\mathbf{A}_{x, y}$ are the values at position $(x, y)$ of correlation matrix $\mathbf{M}$
and resulting attention matrix $\mathbf{A}$ respectively.
The operation of applying softmax on slot dimension normalizes the contributing weights of each slot, hence slots will be distinct from each other and the redundancy among slots will be suppressed. 

In the final step, the resulting attention matrix will be applied to the value features $\mathbf{V}_{g}$ to retrieve related object's information.
This can be formulated as:
\begin{equation}
\begin{split}
    \mathbf{O} &=
    \mathbf{A}^{T} \cdot
    \mathbf{V}_{g}
    ,
    \\
\end{split}
\label{eq:query_end}
\end{equation}
where $\mathbf{O} \in \mathbb{R}^{ L_{q} \times C}$ represents the retrieved object's information.

\noindent\textbf{Panoptic Retriever.}
Panoptic Retriever processes the features with position embedding and panoptic slots of each frame sequentially. Self-attention, Retriever, and Feed Forward Network (FFN) make up the Panoptic Retriever.
Self-attention and FFN are for refining the panoptic slots before and after the association of the panoptic slots with spatial features through Retriever, respectively.
Take one frame $t$ as an example, 
given features $\mathbf{X_t}$ of certain scale, panoptic slots $\mathbf{S}^{i-1}_t$ and position embedding $\mathbf{P}$,
then the procedure of the Panoptic Retriever can be written as:
\begin{equation}
\begin{split}
     \mathbf{\hat S}^{i}_{t} &= \mathbf{S}^{i-1}_{t} + \text{SA}(\mathbf{S}^{i-1}_{t}), \\ 
    \mathbf{\hat S}^{i}_{t} &= \mathbf{\hat S}^{i}_{t} + \text{RE}(\mathbf{\hat S}^{i}_{t}, (\mathbf{X_t} + \mathbf{P}), \mathbf{X_t}), \\ 
    \mathbf{\hat S}^{i}_{t} &= \mathbf{\hat S}^{i}_{t} + \text{FFN}(\mathbf{\hat S}^{i}_{t}), \\
\end{split}
\label{eq:panoptic_retriever}
\end{equation}
where SA, RE refers to Self-attention and Retriever. 
$\mathbf{\hat S}^{i}_{t}\in \mathbb{R}^{L \times C}$ denotes the output panoptic slots refined with spatial information through Panoptic Retriever.
The same operations will be applied to the frame $t-1$, and the output panoptic slots of frame $t-1$ can be denoted as $\mathbf{\hat S}^{i}_{t-1} \in \mathbb{R}^{L \times C}$.
Note that Layer Normalization after summation operation is omitted in the equations of this paper for simplicity.

In the above process, Retriever retrieves the object's information (\eg location, appearance information) from spatial features through associating panoptic slots with every pixel in the spatial features.
Panoptic slots ($\mathbf{\hat S}^{i}_{t}$ or $\mathbf{\hat S}^{i}_{t-1}$) in a single frame act as the $\mathbf{Q}$ in Retriever, 
and $\mathbf{K}, \mathbf{V}$ in Retriever are based on the spatial features of each frame.
Position embedding is added to the spatial features 
to enhance spatial information.
The slots competing mechanism of Retriever facilitate that panoptic slots are mutually exclusive so that only one panoptic object is allocated to one panoptic slot.

\noindent\textbf{Video Retriever.}
The Video Retriever consists of Retriever and FFN.
It aims at relating the panoptic slots across frames and helping these slots corresponding to the same object refine with each other's information.
In this way, the consistency of the panoptic slots describing the same object across frames will be much improved.

Given the output panoptic slots ($\mathbf{\hat S}^{i}_{t}, \mathbf{\hat S}^{i}_{t-1}$) from Panoptic Retriever,
these panoptic slots will be 
concatenated along the slot dimension and 
then forwarded into the Retriever. 
The following FFN is applied to refine the output panoptic slots of Retriever.
And the final refined panoptic slots will be re-distributed to the panoptic slots of the corresponding frame.
The process in Video Retriever can be summarized as follows:
\begin{equation}
\begin{split}
    \mathbf{\hat S}^{i}_{a} &= [\mathbf{\hat S}^{i}_{t}, \mathbf{\hat S}^{i}_{t-1}], \\
    \mathbf{S}^{i}_{a} &= \mathbf{\hat S}^{i}_{a} + \text{RE}(\mathbf{\hat S}^{i}_{a}, \mathbf{\hat S}^{i}_{a}, \mathbf{\hat S}^{i}_{a}), \\
    \mathbf{S}^{i}_{a} &= \mathbf{S}^{i}_{a} + \text{FFN}(\mathbf{S}^{i}_{a}), \\
    [\mathbf{S}^{i}_{t}, \mathbf{S}^{i}_{t-1}] &= \mathbf{\hat S}^{i}_{a} + \mathbf{S}^{i}_{a}, \\
\end{split}
\label{eq:video_retriever}
\end{equation}
where $\mathbf{\hat S}^{i}_{a} \in \mathbb{R}^{2L \times C}$ denotes the concatenated panoptic slots, RE refers to Retriever, [·, ·] indicates the concatenation operation, 
$\mathbf{S}^{i}_{a}$ represents the refined concatenated panoptic slots,
and $\mathbf{S}^{i}_{t}, \mathbf{S}^{i}_{t-1} \in \mathbb{R}^{L \times C}$ denote the spatio-temporally refined
output panoptic slots of frame $t$ and $t-1$, respectively.

Different from the Retriever in Panoptic Retriever, 
the concatenated panoptic slots serve as 
$\mathbf{Q}$, $\mathbf{K}$, and $\mathbf{V}$ in Retriever,
which utilizes the slots competing mechanism to
help make sure 
the unique connection between two specific panoptic slots across frames.
This also paves the way for ID assignment in ID prediction head, which already can achieve good performance without relying on any other information but only the spatio-temporally refined panoptic slots.
Besides, the whole process of Video Retriever is friendly for variational frame numbers due to the parameters are only related to the length of slot vector.

\noindent \textbf{Prediction heads.}
After the above Panoptic Retriever and Video Retriever operations, the panoptic slots will contain the object's information in each frame and slots corresponding to the same object across frames will be as consistent as possible.
To produce classification, mask, and object ID predictions from panoptic slots, we utilize three prediction heads, each consisting of an FFN (two Linear layers) and the respective functional layer.
In the classification head, the FFN is followed by another Linear layer to output the class prediction for each panoptic object.
In the mask head, the mask of the panoptic object is obtained by applying panoptic slots on the feature map through matrix multiplication operation.
In the ID head, the ID of the panoptic object is predicted by calculating the similarity matrix between the panoptic slots in the current frame and previous frames.
Benefitting from the coherent panoptic slots, all prediction heads can provide precise predictions based on the slot information and no complex fusion operations are required.

\section{Experiments}
\subsection{Implementation Details}
\noindent\textbf{Cityscapes-VPS.}
Cityscapes-VPS \cite{kim2020video} is built on top of the validation set of the Cityscapes dataset \cite{cityscapes}. It provides dense panoptic annotations for $3000$ frames, sampling six frames from every $500$ video clips
where a single video clip contains $30$ frames of $1024 \times 2048$ resolution with $19$ classes (11 stuff and 8 things), and instance ID association across frames is also provided.
The split of train, validation, and test sets are $400, 50, 50$ clips respectively.
We report results on its validation set and test set.

\noindent\textbf{VIPER.}
VIPER dataset for the VPS task \cite{kim2020video} is re-formatted based on the synthetic VIPER dataset \cite{richter2017playing} extracted from the GTA-V game engine.
The re-formatted VIPER dataset contains panoptic annotations for $23$ classes ($13$ stuff and $10$ things) on 
184K frames of ego-centric
driving scenes at $1080 \times 1920$ resolution. 
We follow the public train and val
splits as \cite{kim2020video}.
For training, $19$ videos with total $41464$ frames are utilized.
For evaluation, there are total $600$ images, consisting of the first $60$ frames of 10 validation videos from the day scenario.

\noindent\textbf{Evaluation metric.}
The Video Panoptic Quality (VPQ) \cite{kim2020video} is adopted for evaluation.
As the video extension of Image Panoptic Quality (PQ) \cite{kirillov2019panoptic}, 
VPQ is designed 
for evaluating the spatio-temporal
consistency between the predicted and ground truth panoptic video segmentation.
For a video sequence,
denote the temporal window size as $k \in \{ 0,5,10,15 \} $, 
several snippets can be obtained by sliding the window through the video, 
and the calculation of snippet-level IoU, $|$TP$|$, $|$FP$|$ and $|$FN$|$ is performed, accordingly.
At a dataset level, all these values are collected across all predicted videos and 
the dataset-level VPQ$^k$ result is computed for each class and averaged across all classes.
The final VPQ is computed by averaging over different $k$ values.

\noindent\textbf{Training.}
The implementation is based on the MMDetection \cite{chen2019mmdetection} toolbox.
For most experiments, ResNet50 \cite{he2016deep} and FPN \cite{lin2017feature} serve as a backbone network.
Except for ResNet50, we also validate our method on other backbones, such as Swin-B, Swin-L Transformer \cite{liu2021swin}, \etc
Training losses include four image-level losses \cite{wang2021max} (\ie PQ loss, pixel-wise instance discrimination loss, per-pixel mask-ID cross-entropy loss, and semantic segmentation loss) and an identification loss as a video-level loss. 
Corresponding loss weights are empirically set to $3.0, 1.0, 0.3, 0.5, 0.5$ respectively.
As for training data of the Cityscapes-VPS dataset, we find that the temporal annotation is not very consistent, 
hence we mainly utilize the image annotations and generate the simulated videos from these images through random scaling and translating \cite{zhou2020tracking}.

Distributed training with $8$ GPUs is utilized and batch size is set as $1$ for each GPU.
For all datasets, the optimizer, base learning rate, weight decay, and learning rate scheduler are set to AdamW \cite{loshchilov2017decoupled}, $0.0001, 0.0001$ and StepLR respectively. 
For Cityscapes-VPS, we train for 96 epochs and apply lr decay at 64 and 88 epochs. 
Image panoptic segmentation pretraining for backbone and Panoptic Retriever parts are adopted with Mapillary Vistas \cite{neuhold2017mapillary} and Cityscapes train sequences with pseudo labels \cite{chen2020naive}.
For VIPER, we train for 12 epochs and apply lr decay at 8 and 11 epochs. 
Image panoptic segmentation pretraining
is adopted on VIPER dataset.   
The remaining layers, \eg Video Retriever and ID prediction head are initialized by Kaiming initialization.

\noindent\textbf{Inference.}
For Cityscapes-VPS, all $30$ frames of a video are predicted but only $6$ frames with ground truth labels are evaluated.
For VIPER, all frames are predicted and evaluated.
For all datasets, simple mask filtering and removing are adopted. 
In this process, object masks are filtered with a class confidence score of $0.85$ and 
a pixel confidence score of $0.4$ is applied on things masks. 
The things masks with an overlapping ratio greater than $0.03$ and stuff masks with area less than $4096$ 
are removed.

\begin{figure}
\centering
\includegraphics[width=\linewidth]{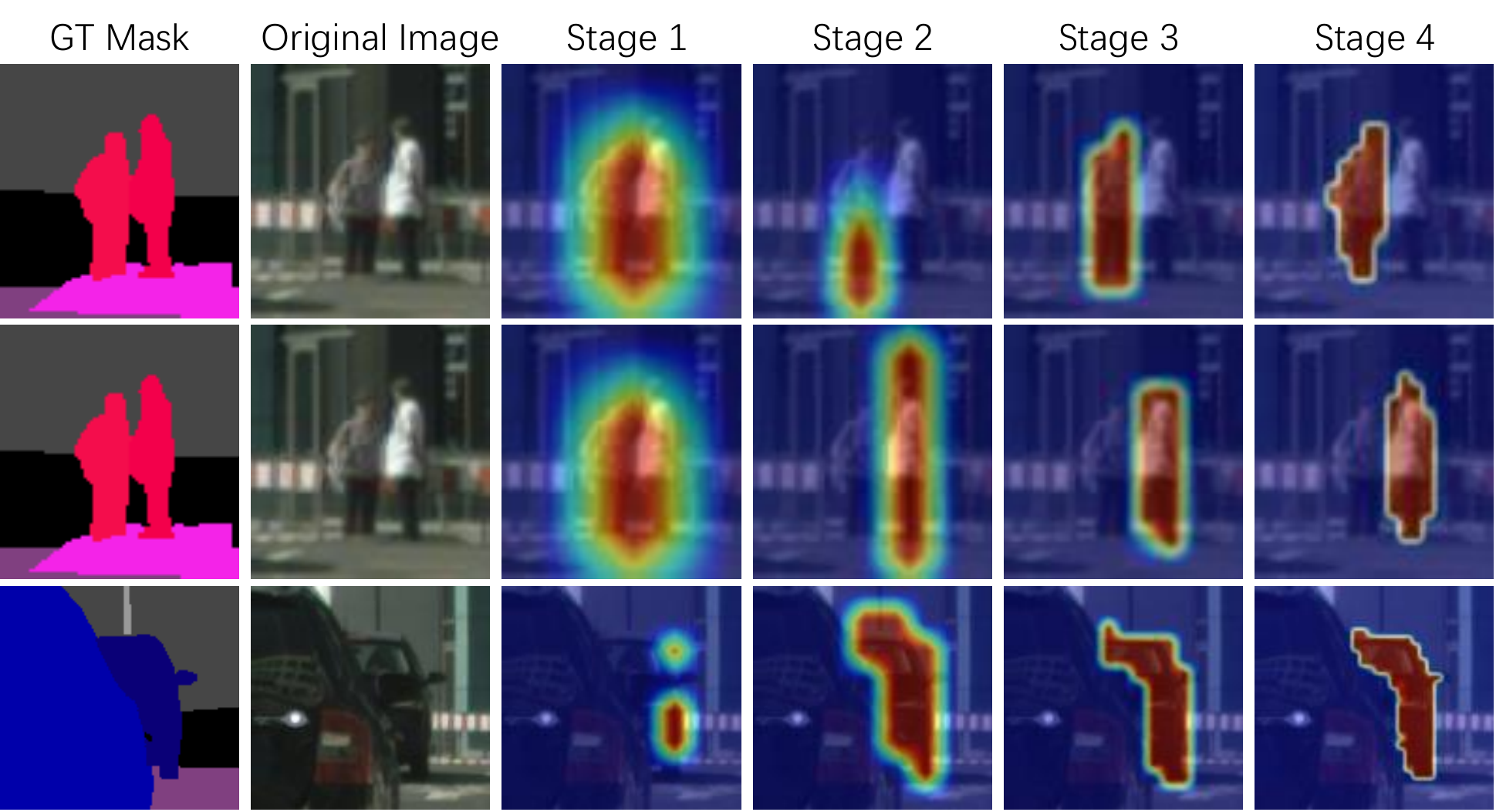}
\caption{\textbf{
Attention maps at different stages.}
Based on the visualization of the CAM (Class Activation Map), We can observe that our panoptic slots only roughly localize objects in early stages and gradually match the object boundaries better and better.
}
\label{fig:vis_multi_stages}
\end{figure}

\begin{figure}
\centering
\includegraphics[width=\linewidth]{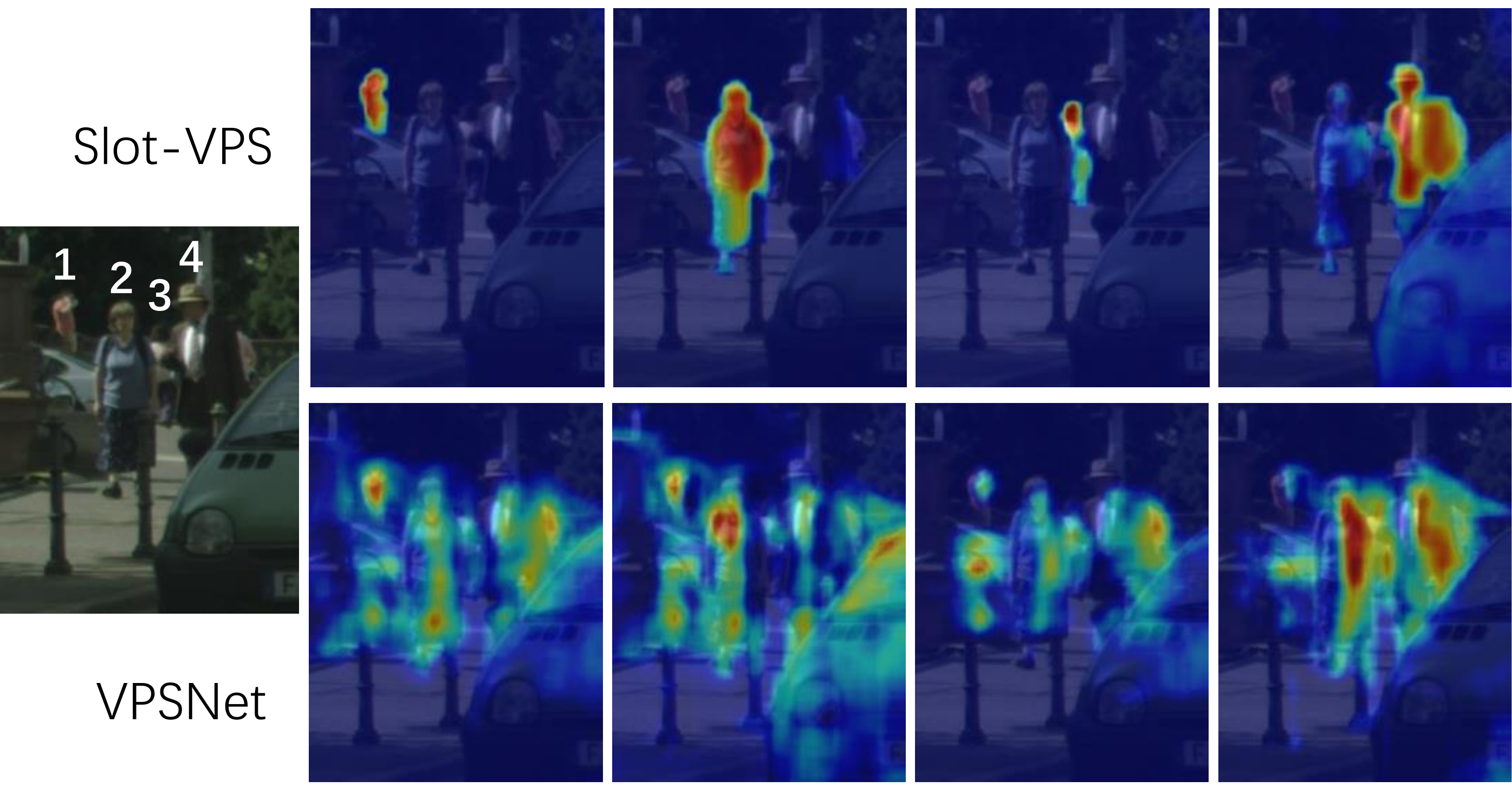}
\caption{
\textbf{Qualitative comparison of feature maps between Slot-VPS (\emph{top}) and VPSNet \cite{kim2020video}(\emph{bottom}).}
 We can find that each of our CAM focuses on a specific object, which verifies that the learned features are object-centric; while the CAMs from \cite{kim2020video} show that their features can not distinguish instances, thus requiring surrogates (\eg bbox, center) to further localize objects. Best viewed when zoomed in.
}
\label{fig:vis_vpsnet_compare}
\end{figure}

\begin{table*}[t]
\begin{minipage}{0.30\textwidth}
\setlength{\tabcolsep}{1.66pt}
\centering
\strut\vspace*{-8pt}
\scalebox{0.86}{
\begin{tabular}{c|c|c}
\Xhline{1.5pt}
 \makecell[c]{Softmax \\ dim}
 & \makecell[c]{\textit{w/}\\Projection}
 & \makecell{PQ / PQ$^{\text{th}}$ / PQ$^{\text{st}}$} \\ \hline
 \rowcolor{Yellow!16} slot 
 & 
 & 60.5 / 48.4 / 69.3 \\
 spatial
 & 
 & 53.5 / 37.5 / 65.1 \\
 slot
 & \checkmark
 & 60.7 / 47.9 / 70.1 \\
\Xhline{1.5pt}
\end{tabular}}
\caption{\textbf{
Variations in Retriever.} Rows rendered in \colorbox{Yellow!16}{yellow} are the final settings.
PQ, PQ$^{\text{th}}$, PQ$^{\text{st}}$ are the averaged scores for all / Things / Stuff classes respectively.
}
\label{table:retriever_comparison}
\end{minipage}
\hspace{0.01\linewidth}
\begin{minipage}{0.37\textwidth}
\setlength{\tabcolsep}{1.66pt}
\centering
\strut\vspace*{-6pt}
\scalebox{0.86}{
\begin{tabular}{c|c|c|c}
\Xhline{1.5pt}
 & \makecell{Encoder} 
 & \makecell[c]{\textit{w/} \\ multi-scale}  
 & \makecell{PQ / PQ$^{\text{th}}$ / PQ$^{\text{st}}$} \\ \hline
\multirow{2}{*}{DETR \cite{carion2020end}} 
 & 
 & \checkmark
 & 38.6 / 27.0 / 47.1 \\
 & \checkmark 
 & \checkmark
 & 56.3 / 44.5 / 64.8 \\ \hline
\multirow{2}{*}{Ours} 
 &
 & 
 & 58.9 / 44.6 / 69.4 \\
  & \cellcolor{Yellow!16}
 & \cellcolor{Yellow!16}\checkmark
 & \cellcolor{Yellow!16}\textbf{60.5 / 48.4 / 69.3} \\ \Xhline{1.5pt}
\end{tabular}
}
\caption{\textbf{
Comparison between Panoptic Retriever and DETR \cite{carion2020end}.} 
``\textit{w/} multi-scale" stands for whether utilizing the multi-scale features.
}
\label{table:SSA_vs_transformer}
\end{minipage}
\hspace{0.01\linewidth}
\begin{minipage}{0.27\textwidth}
\setlength{\tabcolsep}{1.6pt}
\centering
\strut\vspace*{3pt}
\scalebox{0.88}{
\begin{tabular}{c|c}
\Xhline{1.5pt}
  \makecell{Slot number} 
 & \makecell{PQ / PQ$^{\text{th}}$ / PQ$^{\text{st}}$} \\
\hline
 50
 & 57.4 / 42.7 / 68.1 \\
  80
 & 57.6 / 46.4 / 65.7 \\
  \rowcolor{Yellow!16} 100
 & \textbf{60.5 / 48.4 / 69.3} \\
  200
 &  58.7 / 47.6 / 66.7 \\
 300
 & 58.2 / 45.7 / 67.4 \\
\Xhline{1.5pt}
\end{tabular}}
\caption{\textbf{
Comparison between different slot number.}
}
\label{table:slot_number}
\end{minipage}
\end{table*}
\begin{table}
\centering
\scalebox{0.8}{
\begin{tabular}{c|c|c|c|c|c}
\Xhline{1.5pt}
\makecell{FFN \\ Hidden dim}
 & \makecell[c]{scale \\ $/$ 32} 
 & \makecell[c]{scale \\ $/$ 16} 
 & \makecell[c]{scale \\ $/$ 8}  
 & \makecell[c]{scale \\ $/$ 4}
 & \makecell{PQ / PQ$^{\text{th}}$ / PQ$^{\text{st}}$} \\
\hline
\multirow{3}{*}{1024}
 & 1
 & 2
 & 2
 & 2
 & 56.4 / 40.9 / 67.6 \\
 & 2
 & 2
 & 2
 & 2 
 & 58.9 / 44.7 / 69.2 \\
 & 2
 & 3
 & 3
 & 3 
 & 60.1 / 48.3 / 68.7 \\
 \hline
 \multirow{3}{*}{2048}
   & 1
 & 1
 & 1
 & 1
 & 51.6 / 30.1 / 67.2 \\
  & 1
 & 1
 & 1
 & 2
 & 56.1 / 45.4 / 63.8 \\
 & \cellcolor{Yellow!16}1
 & \cellcolor{Yellow!16}2
 & \cellcolor{Yellow!16}2
 & \cellcolor{Yellow!16}2
 & \cellcolor{Yellow!16}\textbf{60.5 / 48.4 / 69.3} \\
\Xhline{1.5pt}
\end{tabular}}
\caption{\textbf{
Comparison between module number variance on multi-scale features.}
}
\label{table:stage_number_channel_size}
\end{table}

\subsection{Ablation Study}
We first analyze the panoptic slots, then study several aspects to discuss the effect of different settings on overall performance.
Unless otherwise specified, experiments are conducted with ResNet50 and FPN on Cityscapes-VPS \textit{val} in this section.

\noindent\textbf{What have panoptic slots learnt?}
To better understand 
what panoptic slots are and what panoptic slots have learnt,
we show the visualization of activation maps on 
the attention maps at different stages in Figure \ref{fig:vis_multi_stages}.
The maps show the most contributing region to a specific panoptic object in the image. 
They are computed with  SEG-GRAD-CAM \cite{vinogradova2020towards}, which is the extension of Grad-CAM \cite{selvaraju2017grad} to semantic segmentation. 
Thanks to the unified end-to-end framework of Slot-VPS, we can compute the loss directly for a predicted instance, and obtain its CAM (Class Activation Map) for any feature or attention map.
As shown in Figure \ref{fig:vis_multi_stages}, 
at early stages, multiple objects may be activated together and the focusing area is large.
With the iterative learning process, other irrelevant information in the scene is suppressed 
and the target object becomes more clearly distinguishable.

\noindent\textbf{Retriever.}
There are two main differences between our Retriever and the classic attention \cite{vaswani2017attention}.
The first one is the dimension on which the softmax operation is performed.
We apply softmax to the slot dimension (\ie Query's dimension) instead of the dimension of spatial size (height $\times$ width) of feature maps (\ie Key's dimension).
Applying softmax to the spatial size dimension enhances the pixel-level discriminability so that the location and appearance information of objects can be obtained from features but object-level relations are not explored,
while applying softmax on the slot dimension can facilitate the object-level competition so that the discriminability of objects can be enhanced.
The second one is whether there is a projection layer after applying the attention matrix on the Value features.
We conduct ablation studies on these two factors of Retriever on Image Panoptic Segmentation task.
As shown in Table \ref{table:retriever_comparison}, we can observe that changing the softmax's applying dimension from
slot to spatial size
degrades the performance a lot, which validates that the slots competing mechanism at object-level is effective.
Adding the final projection layer only brings minor performance gain, so we do not keep this layer for overall efficiency.

\begin{table*}
\centering
\scalebox{0.8}{
\begin{tabular}{c|c|c|c|c|c|c|c}
\Xhline{1.5pt}
    \multirow{2}{*}{} &
    \multirow{2}{*}{Methods} &
    \multicolumn{4}{c|}{Temporal window size $k$}  &
    \multirow{2}{*}{VPQ/VPQ$^{\text{th}}$/VPQ$^{\text{st}}$} &
    \multirow{2}{*}{FPS} \\ 
    \cline{3-6}
    & & $k$=0 & $k$=5 & $k$=10 & $k$=15 & &  \\ \hline
    \multirow{3}{*}{\makecell{Cityscapes-VPS \\ val }} &
     VPSNet \cite{kim2020video} 
     & 64.5 / 58.1 / 69.1
     &  57.4 / 45.2 / 66.4
     & 54.1 / 39.5 / 64.7
     & 52.2 / 36.0 / 64.0
     & 57.0 / 44.7 / 66.0
     & 1.3 \\
      & SiamTrack \cite{woo2021learning}
      & 64.6 / 58.3 / 69.1
    & 57.6 / 45.6 / 66.6
    & 54.2 / 39.2 / 65.2
     & 52.7 / 36.7 / 64.6
     & 57.3 / 44.7 / 66.4
     & 4.6 \\ \cline{2-8}
      & \multirow{1}{*}{Ours}
      & \cellcolor{Yellow!16}65.7 / 57.9 / 71.4
    & \cellcolor{Yellow!16}60.0 / 47.7 / 68.9
    & \cellcolor{Yellow!16}57.8 / 44.4 / 67.6
     & \cellcolor{Yellow!16}55.5 / 40.2 / 66.7
     & \cellcolor{Yellow!16}\textbf{59.7 / 47.5 / 68.6} 
     & \cellcolor{Yellow!16}\textbf{4.6} \\ \Xhline{1.5pt}
     \multirow{3}{*}{\makecell{Cityscapes-VPS \\ test }} &
     VPSNet \cite{kim2020video}
     & 64.2 / 59.0 / 67.7
     &  57.9 / 46.5 / 65.1
     & 54.8 / 41.1 / 63.4
     & 52.6 / 36.5 / 62.9
     & 57.4 / 45.8 / 64.8 
     & 1.3 \\ 
      & SiamTrack \cite{woo2021learning}
      & 63.8 / 59.4 / 66.6
    & 58.2 / 47.2 / 65.9
    & 56.0 / 43.2 / 64.4
     & 54.7 / 40.2 / 63.2
     & 57.8 / 47.5 / 65.0
     & 4.6 \\
     \cline{2-8}
      & Ours
      & \cellcolor{Yellow!16}66.4 / 58.8 / 71.2
    & \cellcolor{Yellow!16}60.9 / 48.8 / 68.5
    & \cellcolor{Yellow!16}57.5 / 44.2 / 65.9
     & \cellcolor{Yellow!16}55.8 / 40.8 / 65.4
     & \cellcolor{Yellow!16}\textbf{60.1 / 48.2 / 67.8}
     & \cellcolor{Yellow!16}\textbf{4.6} \\  
    \Xhline{1.5pt}
\end{tabular}}
\caption{\textbf{
Comparison to the state-of-the-art with ResNet50-FPN on Cityscapes-VPS \textit{val} and \textit{test}.} 
VPQ, VPQ$^{\text{th}}$, VPQ$^{\text{st}}$ are the averaged scores for all / Things / Stuff classes respectively.
}
\label{table:sota}
\end{table*}

\begin{table*}
\centering
\scalebox{0.88}{
\begin{tabular}{c|c|c|c|c|c|c|c|c|c}
\Xhline{1.5pt}
 & \makecell[c]{DenseContext} 
 & \makecell[c]{AutoAug}  
 & \makecell[c]{RFP}
 & \makecell[c]{SSL}
 & \makecell[c]{TTA}
 & \makecell[c]{Backbone}
 & \makecell[c]{Image model \\ M-adds (B)}
 & \makecell[c]{val \\ VPQ}
 & \makecell[c]{test \\ VPQ}\\
\hline
  \multirow{3}{*}{ViP-Deeplab \cite{qiao2020vip}}
 & \checkmark
 & 
 & 
 & 
 & 
 & \multirow{3}{*}{WR-41}
 & \multirow{3}{*}{3246} 
 & 60.9
 & - \\
 & \checkmark
 & \checkmark
 & \checkmark
 & 
 & 
 & 
 & 
 & 61.9
 & - \\
 & \checkmark
 & \checkmark
 & \checkmark
 & \checkmark
 & \checkmark
 & 
 & 
 & 63.1 
 & 62.5 \\
 \hline
 \rowcolor{Yellow!16} 
 & 
 & 
 &  
 & 
 & 
 & 
 & 
 & 62.2
 & 61.6 \\
 \rowcolor{Yellow!16}
 \multirow{-2}{*}{Ours}
 & 
 & 
 &  
 & 
 & \checkmark
 & \multirow{-2}{*}{Swin-L-FPN}
 & \multirow{-2}{*}{1917}
 & \textbf{63.7}
 & \textbf{63.3} \\
\Xhline{1.5pt}
\end{tabular}}
\caption{\textbf{
Comparison to the state-of-the-art \cite{qiao2020vip} on Cityscapes-VPS \textit{val} and \textit{test}.
}
We list the tricks applied in both methods including DenseContext\cite{qiao2020vip}, AutoAugment \cite{cubuk2019autoaugment}, Recursive Feature Pyramid (RFP) \cite{qiao2021detectors}, Semi-Supervised Learning (SSL) with pseudo labels \cite{chen2020naive} and Test Time Augmentation (TTA) \cite{qiao2020vip}.
}
\label{table:vip_compare}
\end{table*}
\newcommand{\gain}[1]{\textcolor{Green}{(+{#1})}}
\begin{table}
\setlength{\tabcolsep}{5.66pt}
\centering
\begin{tabular}{c|c|c|c}
\Xhline{1.5pt}
\makecell[c]{Methods} 
 & \makecell[c]{Backbone} 
 & \makecell[c]{VPQ} 
 & \makecell[c]{FPS}  \\
 \hline
  SiamTrack \cite{woo2021learning}
 & ResNet50-FPN
 & 50.2 
 & 5.1 \\
   VPSNet \cite{kim2020video}
 & ResNet50-FPN
 & 51.9 \gain{1.7}
 & 1.0 \\ 
 \hline
 \multirow{2}{*}{Ours} 
 & \cellcolor{Yellow!16}ResNet50-FPN 
 & \cellcolor{Yellow!16}53.2 \gain{3.0}
 & \cellcolor{Yellow!16}4.2 \\
 & \cellcolor{Yellow!16}Swin-L-FPN 
 & \cellcolor{Yellow!16}\textbf{56.2} \gain{6.0}
 & \cellcolor{Yellow!16}2.0 \\
\Xhline{1.5pt}
\end{tabular}
\caption{\textbf{
Comparison to the state-of-the-art on VIPER \textit{val} for VPS task.} 
}
\label{table:viper_compare}
\end{table}
\noindent\textbf{Panoptic Retriever \textit{vs.} Transformers in DETR.}
Our Panoptic Retriever is inspired by the decoder structure in the transformer (\eg DETR \cite{carion2020end}). 
The transformers in DETR \cite{carion2020end} consists of encoder, decoder, and panoptic mask head. Concretely, there are three key differences between the Panoptic Retriever and the transformer structure in DETR:

\noindent(1) 
The attention mechanism in DETR applies the softmax along the spatial dimension, which only discriminates different pixels instead of competing among objects.
Our Panoptic Retriever applies softmax on the slot dimension to encourage the competition between slots so that the information of each slot becomes mutually exclusive. Table \ref{table:retriever_comparison} validates the efficacy of such a competition mechanism.

\noindent(2) 
DETR highly relies on the transformer encoders on top of the backbone. The transformer encoder brings great computational cost and greatly slows down the entire framework.
Without the help of the encoder, our Panoptic Retriever already achieves good performance and can be applied efficiently.

\noindent(3)
DETR only uses the low-resolution feature maps in the encoder and fuses multi-scale features in the extra panoptic mask head, which could be harmful to the modeling of small objects. Instead, we feed and fuse multi-scale features into Panoptic Retriever without requiring additional heads. 

As shown in the first two rows of Table \ref{table:SSA_vs_transformer}, removing the encoder of the transformer will lead to great degradation, while our network can perform much better
without the help of the encoder ($4$th row).
Even without the use of multi-scale features (the $3$rd row), we can still outperform the DETR with six transformer encoders by $2.6$ PQ.
Note that 
utilizing multi-scale features mainly improves things' performance.

\paragraph{Slot number.}
In this part, we explore the effect of different slots numbers as shown in Table \ref{table:slot_number}.
As it can be observed, since a slot is a structure designed to take full responsibility for a single object, it is desirable to set the slot number close to the number of available panoptic objects in a scene (\eg 100 in this setting).
If the slot number is too small, the multiple objects are more likely to be assigned to the same panoptic slot, resulting in confusion between panoptic objects.
Some objects will be missed due to insufficient slots.
In contrast, if the slot number is too large,
a single object might spread into multiple slots as fragments during the competition, which generates less reliable slots compared to the ideal case of each slot responsible for a single object.

\noindent\textbf{Module number in each stage.}
To sufficiently refine the slots with spatio-temporal coherent information, we apply the VPR module multiple times with multi-scale features.
We survey the effect of different module numbers on each scale under two settings,
setting the hidden dim in FFN of model to $1024$ and $2048$ respectively.
As shown in Table \ref{table:stage_number_channel_size},
the performance is improved with the increase of per-scale module number for both settings.
However, this will bring extra computation complexity.
By default, we empirically set the module number of four scales features to $1, 2, 2, 2$ respectively.
As in Figure \ref{fig:comparison_fps}, such a setting can achieve a better balance between the performance and latency 
($59.7$ VPQ with $4.6$ FPS on Cityscapes-VPS \textit{val} set). 

\noindent\textbf{Video Retriever.}
Previous works exploit the temporal information 
through estimating optical flow \cite{ilg2017flownet}
or applying attention mechanisms \cite{wang2018non}.
However, these techniques are mostly applied to the extracted features, which contain all objects' information in the scene.
This also leads to the requirement for extra surrogates (\eg bbox, center) to localize objects from features.
In comparison, our Video Retriever is directly applied to the object-centric representations, which will  eliminate the effect of irrelevant background noises and benefit the object-level mutual refinement.
As shown in Figure \ref{fig:vis_vpsnet_compare}, 
take VPSNet \cite{kim2020video}
as an example, 
thanks to object-centric learning,
our feature maps are more object-centric while the feature maps of \cite{kim2020video} are lack of specific objects' information.
Experiment shows that there is $1.5$ VPQ drop when Video Retriever is removed. 

\subsection{Comparison with the State-of-the-art.}
The VPS results on Cityscapes-VPS are shown in Table \ref{table:sota}.
With the same backbone (ResNet50-FPN), our model outperforms the SiamTrack \cite{woo2021learning} by $2.4, 2.3$ VPQ on Cityscapes-VPS \textit{val} and \textit{test} respectively with the fastest inference speed.
As exhibited in Table \ref{table:vip_compare}, our method can also outperform the state-of-the-art
\cite{qiao2020vip} by $0.6, 0.8$ VPQ respectively
with much less computation and fewer tricks applied,
which indicates that our framework could be further boosted with similar tweaks in the future.
The VPS results on the VIPER dataset are shown in Table \ref{table:viper_compare}.
With the same backbone (ResNet50-FPN), our model outperforms the state-of-the-art \cite{kim2020video} by $1.3$ VPQ on the val set with $4 \times$ FPS as \cite{kim2020video}.
With the larger backbone (Swin-L and FPN), our performance can be further improved by $3.0$ VPQ while achieving $2 \times$ faster as \cite{kim2020video}.
Qualitative visualizations will be shown in the complementary materials.

\section{Conclusion}
In this paper, to alleviate the disadvantages of the decomposed pipeline for the VPS task, we introduce a fully unified end-to-end framework, Slot-VPS, based on object-centric representation learning. 
Panoptic objects (including things and stuff) in the video are represented with a unified representation called panoptic slots.
The proposed Video Panoptic Retriever (VPR) retrieves and encodes the spatio-temporal information of objects in the video into the panoptic slots.
Finally, the spatio-temporal coherent panoptic slots can be utilized for directly predicting the class, mask, and object ID of panoptic objects in the video. 
Experimental results on two datasets of the VPS task validate the effectiveness of our method.

\noindent\textbf{Limitation and broader impact.}
(1) 
Slot-VPS currently predicts the object ID based on the correlations of panoptic slots across frames. Advanced architecture, loss, and regularization technologies may be explored to improve VPS performance without an ID head.
(2) Slot-VPS unifies the video panoptic segmentation in terms of representations, however, it does not fully unify the entire training pipeline since the learning targets are still separated and individual losses and manually tuned loss weights are needed. 
This problem is challenging but maybe potentially solved by designing a new unified loss to the whole VPS task.
(3) Current metric does not fully consider the severity of prediction error, which is also an interesting research direction.

{\small
\bibliographystyle{ieee_fullname}
\bibliography{egbib}
}

\appendix
\renewcommand\thefigure{S\arabic{figure}}
\setcounter{figure}{0}

\begin{figure*}[t]
\centering
\includegraphics[width=\textwidth]{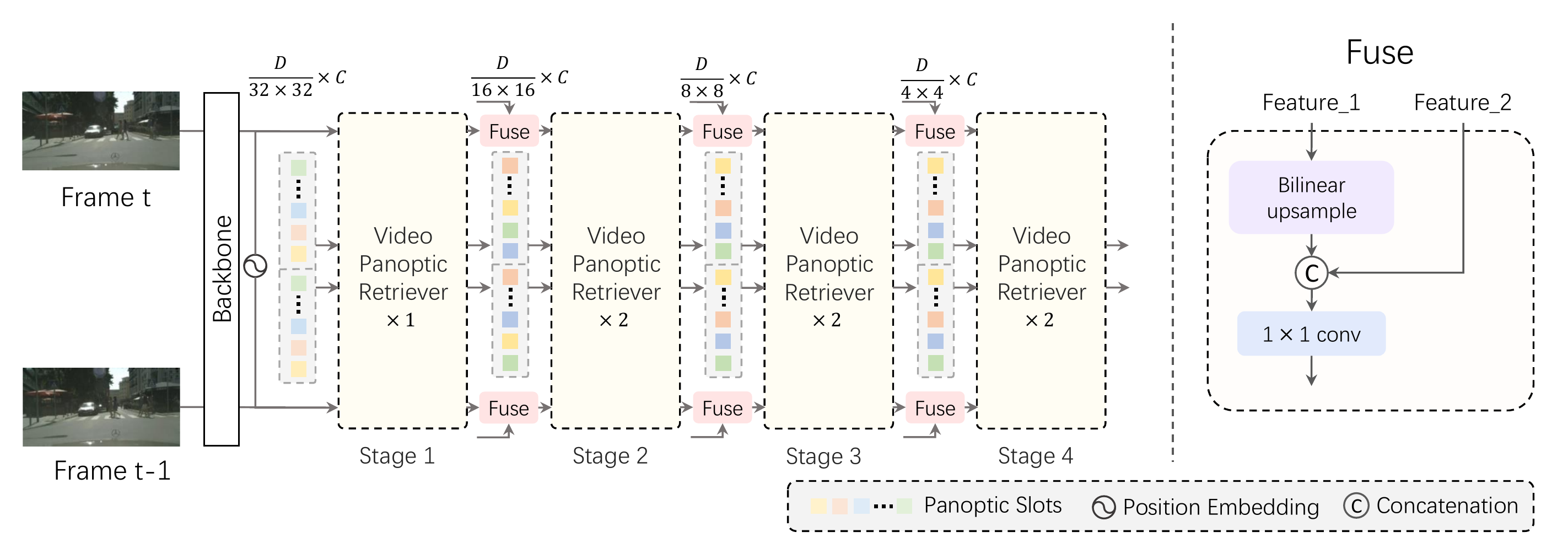}
\caption{\textbf{Multi-scale feature utilization in the Slot-VPS.}
Take two frames (t and t-1) as an example, 
four stages of Video Panoptic Retriever (VPR), consisting of $1, 2, 2, 2$ VPR modules respectively, are consecutively applied on four scales of multi-scale features extracted from the backbone.
In each stage, VPR takes the features with its position embedding and panoptic slots as input and output the spatio-temporal coherent panoptic slots.
Across stages, Fuse module is applied to generate fused features for later stage.
Note that position embedding for later stages are omitted for brevity. 
$D, C$ refer to the spatial size (height $\times$ width) and the number of channels of feature maps respectively.
}
\label{fig:multi_scale}
\end{figure*}

\section{Multi-scale features utilization.}
To exploit sufficient spatio-temporal information, in Slot-VPS, VPR (Video Panoptic Retriever) is stacked for multiple stages and employed with multi-scale features.
As shown in Figure \ref{fig:multi_scale}, each stage, consisting of multiple VPR modules, is applied on features of certain scale, and features of different scales are fused through Fuse module across stages.
Sinusoidal position embedding \cite{carion2020end} is generated based on the input features of each stage.
In Fuse module,
the lower-resolution feature map is up-sampled, concatenated with the higher-resolution feature map, and fused via a $1\times1$ convolutional layer.
The fused feature map will be fed into the subsequent 
stage.

With this multi-scale learning strategy, the difficulty of learning the unified slot representations is mitigated and the ability of handling multi-scale objects is improved.

\section{Difference with Slot Attention.}
The difference between Slot Attention \cite{locatello2020object} and VPR mainly lies in the prior assumption and the application scenarios. 
Slot Attention is based on the assumption of ideal normal distribution,
hence it mainly applies to the the synthetic datasets.
Concretely, the initial slots are randomly sampled from a normal distribution where the mean and variance of this distribution are learnable parameters. 
Although this allows the Slot Attention to generalize to a different number of slots at test time,
this also limits the Slot Attention that it only learns the whole distribution of objects in the scene instead of the detailed representations of each object.
Therefore, the application of Slot Attention is usually the synthetic image datasets where the objects in the image are synthetic geometries.
Differently, our panoptic slots, 
a set of learnable parameters,
uniformly represent all panoptic objects (including things and stuff) in the video.
Each panoptic slot corresponds to a panoptic object in the video and VPR encodes the spatio-temporal objects' information into panoptic slots.
Compared with learning the distribution of objects, our object-level representation learning can handle more complex real-world scenarios which contain a variety of objects.

\section{Visualization of result comparison on Cityscapes-VPS and VIPER.}
\begin{figure*}
\centering
\includegraphics[width=\textwidth]{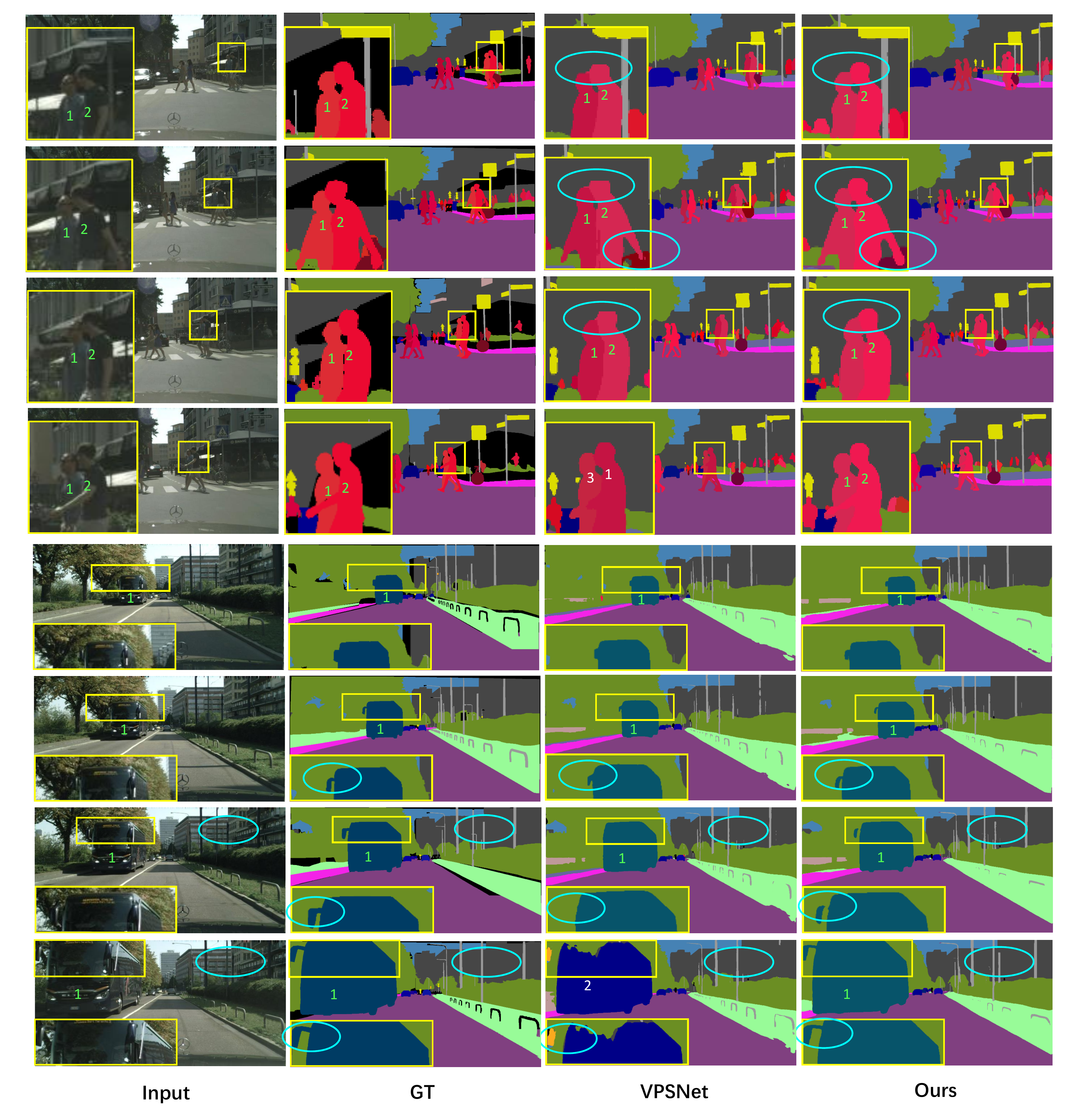}
\caption{\textbf{Visualization of result comparison on Cityscapes-VPS.}
From left to right: input frame, GT annotation, VPSNet predictions, and our predictions.
Each case contains four consecutive frames (from top to bottom) from a video.
Key regions are cropped in yellow rectangles. The matched instances are tagged with the same number across frames. Green and white numbers represent the consistent and inconsistent ID predictions respectively. 
Note that the bus in the fourth frame of bottom case is mistakenly classified as car by VPSNet.
In comparison, our results have better temporal consistency even though people are very close to each other or the bus varies greatly in size.
Moreover, as emphasized by the blue ovals, our segmentation masks have richer details (\eg the head and hand of person in top case, the bus mirror in bottom case). 
Best viewed in color.}
\label{fig:supp_result_0}
\end{figure*}

\begin{figure*}[t]
\centering
\includegraphics[width=\textwidth]{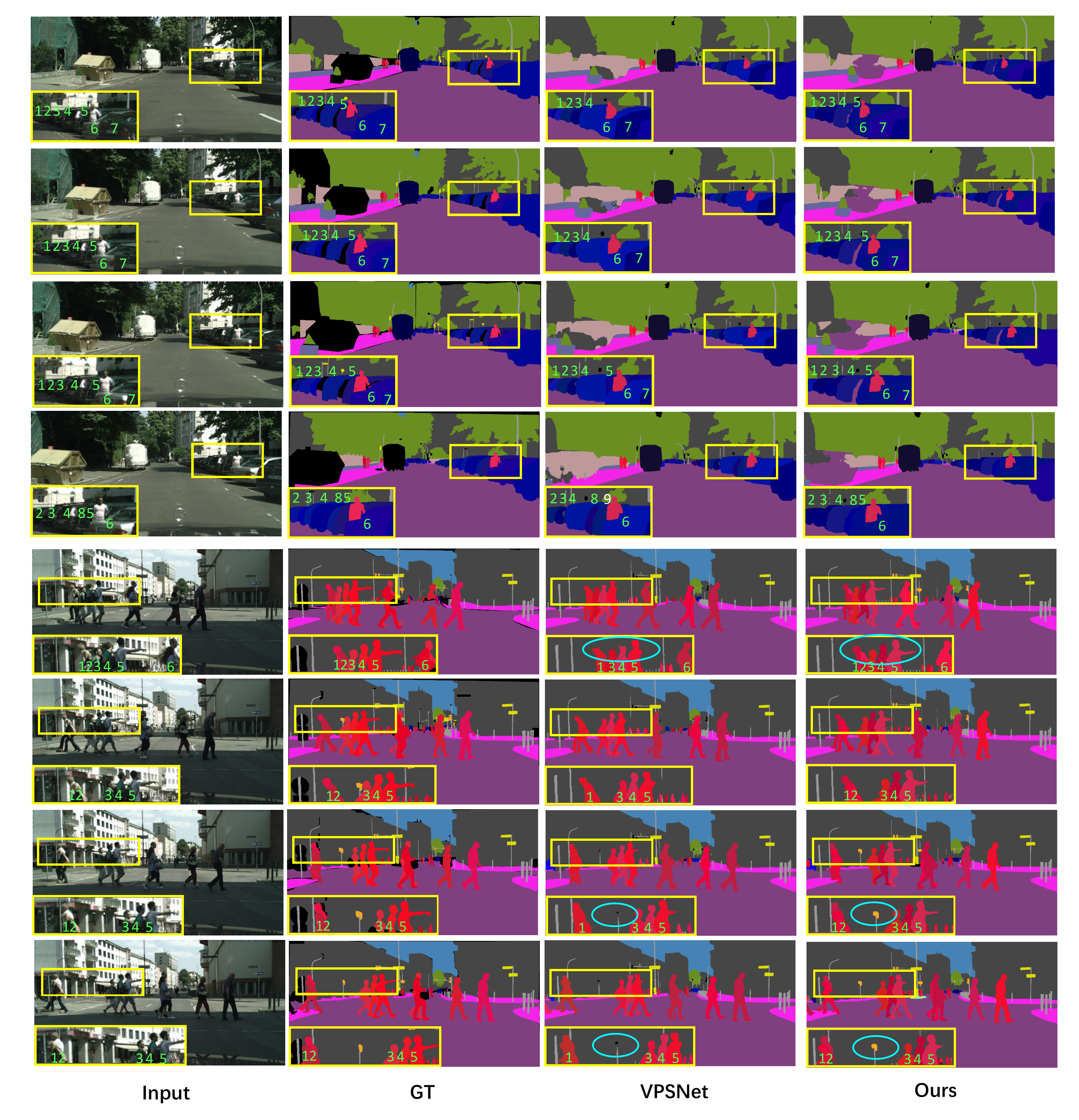}
\caption{\textbf{Visualization of result comparison on Cityscapes-VPS.}
From left to right: input frame, GT annotation, VPSNet predictions, and our predictions.
Each case contains four consecutive frames (from top to bottom) from a video.
Key regions are cropped in yellow rectangles. The matched instances are tagged with the same number across frames. Green and white numbers represent the consistent and inconsistent ID predictions respectively. 
For this kind of dense situations, VPSNet easily missed or mistakenly assigned IDs to objects (\eg the car with ID $5$ in the top case is correctly segmented and identified only in the third frame, the person with ID $2$ in the bottom case is missed for all frames.).
While our predictions are always consistent across all frames.
Moreover, as emphasized by the blue ovals, our segmentation masks have richer details (\eg the head and hand of person, the traffic sign in bottom case). 
Best viewed in color.}
\label{fig:supp_result_1}
\end{figure*}

\begin{figure*}[t]
\centering
\includegraphics[width=\textwidth]{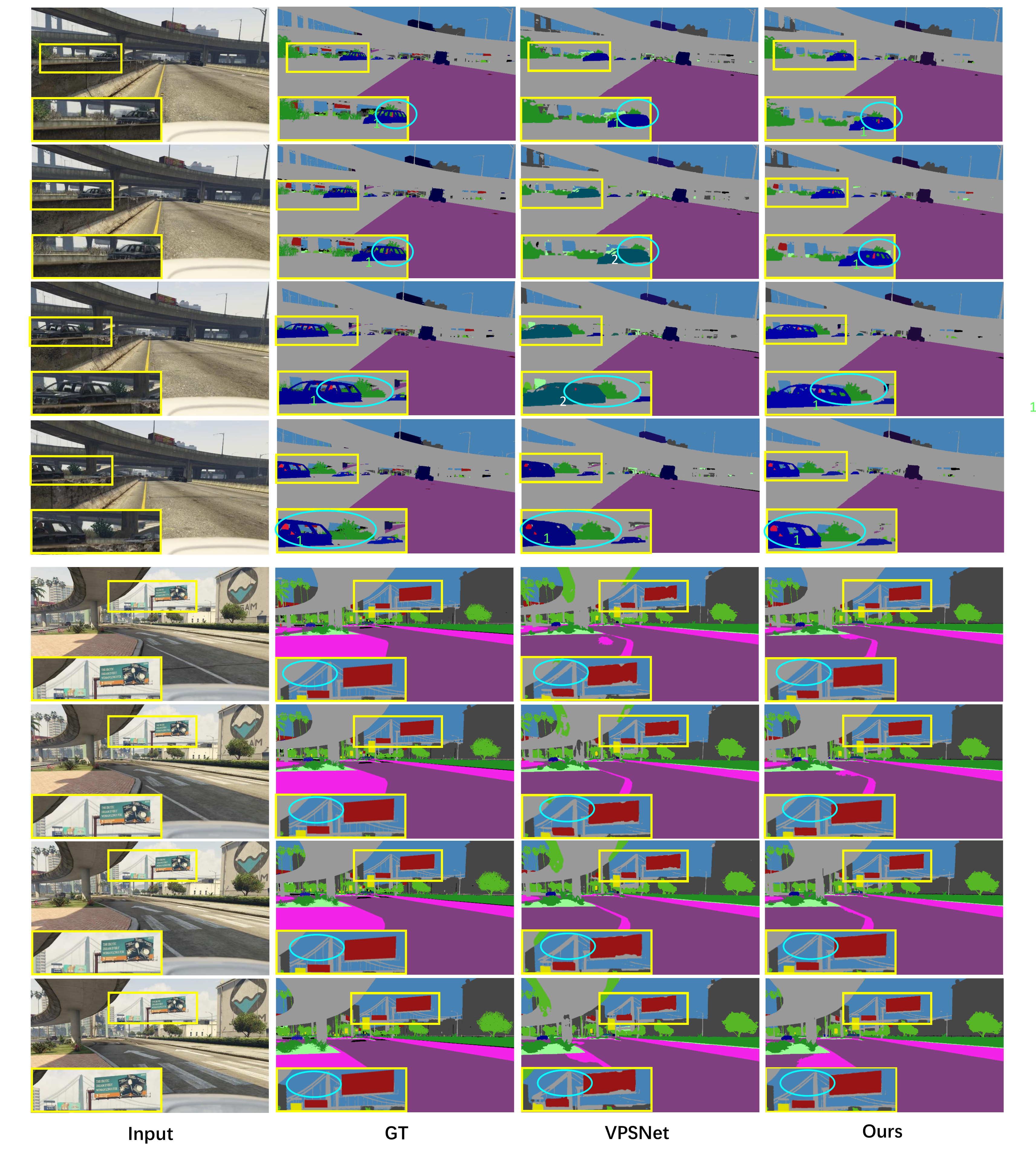}
\caption{\textbf{Visualization of result comparison on VIPER.}
From left to right: input frame, GT annotation, VPSNet predictions, and our predictions.
Each case contains four consecutive frames (from top to bottom) from a video.
Key regions are cropped in yellow rectangles. The matched instances are tagged with the same number across frames. Green and white numbers represent the consistent and inconsistent ID predictions respectively. 
As emphasized by the blue ovals, our segmentation masks have richer details (\eg the person in the car in the top case, the bridge cable in bottom case). 
Best viewed in color.}
\label{fig:supp_result_2}

\end{figure*}

\begin{figure*}[t]
\centering
\includegraphics[width=\textwidth]{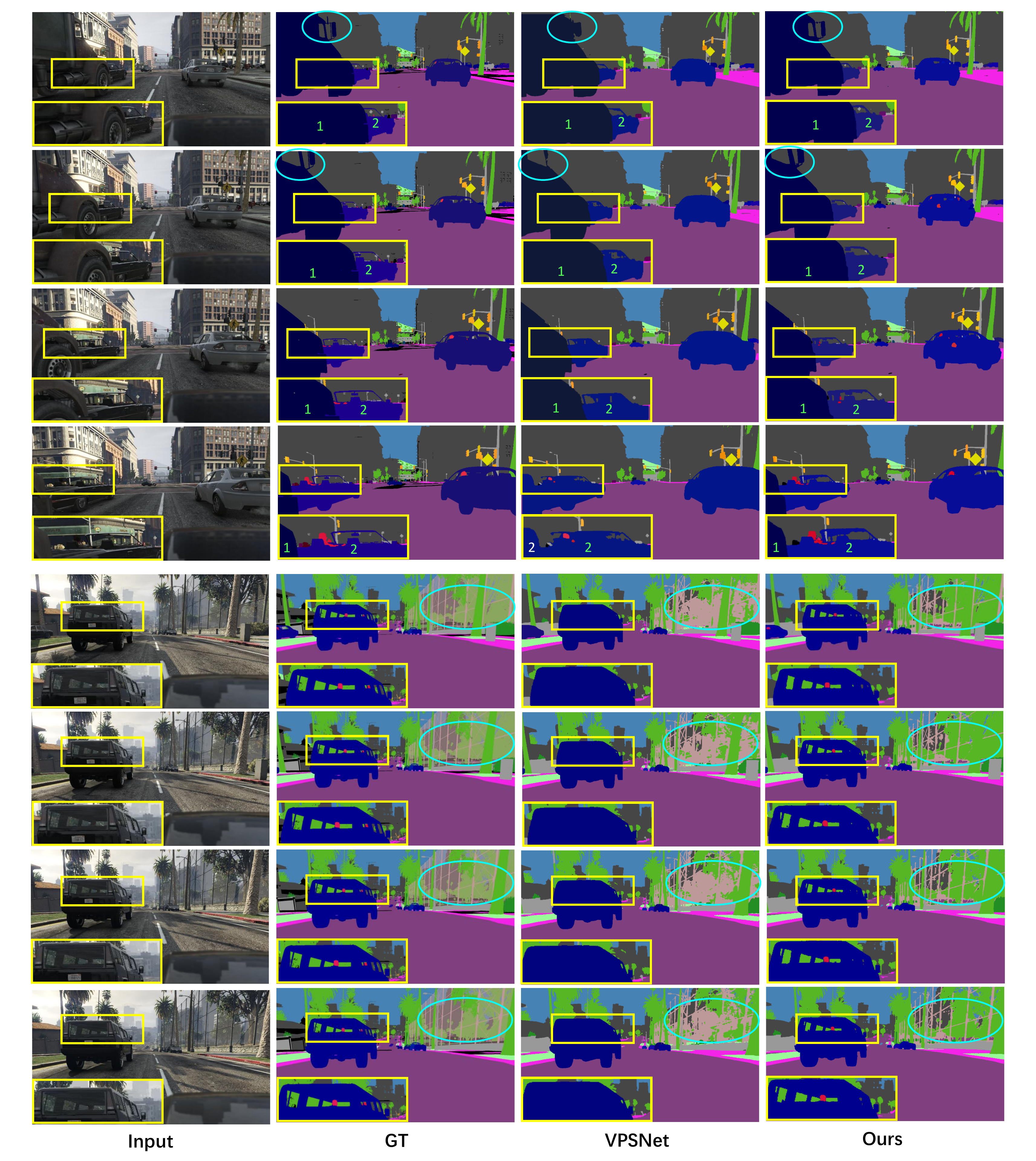}
\caption{\textbf{Visualization of result comparison on VIPER.}
From left to right: input frame, GT annotation, VPSNet predictions, and our predictions.
Each case contains four consecutive frames (from top to bottom) from a video.
Key regions are cropped in yellow rectangles. The matched instances are tagged with the same number across frames. Green and white numbers represent the consistent and inconsistent ID predictions respectively. 
As emphasized by the blue ovals, our segmentation masks have richer details (\eg the car mirror in the top case, the fence in bottom case).
Note that VPSNet can barely segment the details inside cars in these cases but our predictions have sufficient details.
Best viewed in color.}
\label{fig:supp_result_3}
\end{figure*}

Result comparison between VPSNet \cite{kim2020video} and the proposed Slot-VPS on Cityscapes-VPS and VIPER are visualized in Figure \ref{fig:supp_result_0}, Figure \ref{fig:supp_result_1}, Figure \ref{fig:supp_result_2} and Figure \ref{fig:supp_result_3}.
It validates that our method can handle objects with different scales, achieve richer details in single frame and better temporal consistency across frames.


\end{document}